\documentclass[11pt,a4paper]{article}
\usepackage[hyperref]{acl2020}
\usepackage{times}
\usepackage{latexsym}
\usepackage{amsmath}
\usepackage{amssymb}
\usepackage{graphicx}

\usepackage{microtype}

\aclfinalcopy 

\newtheorem{theorem}{Definition}
\newcommand{\norm}[1]{\left\lVert#1\right\rVert}

\title{Persistence Homology of TEDtalk: Do Sentence Embeddings Have a Topological Shape?}

\author{Shouman Das \\
  Univerisity of Rochester \\
  \\\And
  Syed A. Haque \\
    Northeastern University\\
  \\\And
  Md. Iftekhar Tanveer \\
  Comcast Applied Ai Research \\
  }

\date{}

\begin{document}
\maketitle
\begin{abstract}
\emph{Topological data analysis} (TDA) has recently emerged as a new technique to extract meaningful discriminitve features from high dimensional data. In this paper, we investigate the possibility of applying TDA to improve the classification accuracy of public speaking rating. We calculated \emph{persistence image vectors} for the sentence embeddings of TEDtalk data and feed this vectors as additional inputs to our machine learning models. We have found a negative result that this topological information does not improve the model accuracy significantly. In some cases, it makes the accuracy slightly worse than the original one. From our results, we could not conclude that the topological shapes of the sentence embeddings can help us train a better model for public speaking rating. 
\end{abstract}

\section{Introduction}
Storytelling is a major activity through which we  socialize with each other. Using storytelling we teach each other, we influence each other, we pass our values from one generation to another~\cite{pratchett2003science,campbell2008hero,gottschall2012storytelling}. Effective storytelling is a key element for spreading ideas on religion, political issues and even science. Although effective storytelling is quite complex and daunting task for the uninitiated, time to time it has been conjectured that there are some kind of common shapes or arcs that dictates those stories. Prior studies such as \cite{reagan2016emotional} found that the stories have six core emotional arcs by analyzing 1327 stories from project Gutenberg collection. Famous writer Kurt Vonnegut Jr. propsed a fundamental idea that stories have shapes that can be drawn in a graph paper \cite{vonnegut1999palm}. In this vain we try to see whether there is really a topological shape that identifies the storytelling in a public speaking set-up. To explore that idea we used scripts of the TEDTalks, one of the largest available public speaking dataset where there are influential videos from expert speakers conveying ideas on education, business, science through effective storytelling in the public setting.

Our objective in this study is to find the  geometric shapes of public speeches using TDA  and validate if this has any predictive power. As a baseline we use the doc2vec representation of the scripts to perform a classification task. Then represent a public talk as a collection of it's sentences (\emph{sentence cloud} of a talk) from which we extract the topological shape of a talk. We then perform the same classification task with this topological features as an additional input. From our experiments, we cannot categorically decide if the underlying topological shape of the public talk can help us improve the rating.

\section{Related Work}

TDA provides a novel and powerful set of techniques which has been used in data mining, pattern recognition and various machine learning tasks i.e  3d shape detection~\cite{chazal2009gromov}, image processing ~\cite{carlsson2009topology, chung2018topological}, medical biology~\cite{li2015identification,nicolau2011topology}, network analysis~\cite{de2007coverage, carstens2013persistent}, nanotechnology~\cite{nakamura2015persistent}, time series analysis~\cite{umeda2019topological}, graph classification~\cite{hofer2017deep} etc. There have also been some prior research about applying TDA for NLP tasks. \citet{zhu2013persistent} suggested a text representation method by applying persistence homology to the similarity filtration of text data. TDA has been applied for sentiment analysis~\cite{doshi2018movie}, text classification~\citet{gholizadeh2018topological}. This prior work suggests that topological features could be useful for various NLP tasks. However, \citet{michel2017does} found a negative result that text classification accuracy doesn't improve much if one extract the topological features from the \emph{word embedding}. 

\section{Background and Theory}

Due to space constraints, we will avoid explaining the rigorous definitions of persistence homology but give an intuitive explanation of topological features of data. Interested readers are encouraged to see the pioneering works of~\citet{edelsbrunner2000topological},\citet{zomorodian2005computing}. 
\subsection{Persistence Diagram (PD)}
In general, a set of points $A\in \mathbb{R}^d$ (point cloud) can be considered as a sample of points from some underlying topological space $X$. Topological features resembles the number of holes (1-dimensional), void (2-dimensional) present in that topological space $X$. For example, a circle $S^1$ has one 1-dimensional hole, a 2-d sphere $S^2$ has zero 1-dimensional hole, but one 2-dimensional hole (the void inside), the surface of a donut (torus) has two 1-dimensional holes, one 2-dimensional hole. This idea can be generalized to higher dimensions. Now, given a set of points (point clouds) we first create a nested sequence of topological space (Vietoris-Rips filtration, see~\cite{chazal2017introduction} for detail).  Then we calculate the persistence homology of this space which captures the information about the aforementioned topological features (holes, void etc).
By calculating persistence homology of the Vietoris-Rips filtration, we get the persistence diagram (PD) of the data. Intuitively, PD is a set of 2-dimensional points $(b_i,d_i)_{i\in I}$ which represents the birth and death time of a topological feature (for example, a 1-dimensional hole appearing at time $b_i$, disappearing at time $d_i$).    

\subsection{Persistence Image Vector}\label{piv}
Using PDs directly to downstream machine learning task is not straightforward. Several methods have been proposed by algebraic topologists to represent these PDs as vectors which is stable under a suitable distance metric. Some notable examples of these methods include persistence barcodes~\cite{ghrist2008barcodes}, persistence landscape~\cite{bubenik2015statistical} etc. We follow the method of calculating persistence image vector (PIV) which was first introduced by \citet{adams2017persistence}.  
\begin{theorem}
The $p$-Wasserstein metric is defined between two persistence diagrams ${D_1}$ and ${D_2}$ as 
$$
W_p(D_1,D_2) = \inf_{\alpha:D_1\to D_2}\sum_{p\in D_1}\norm{(p-\alpha(p)}^p_\infty
$$
where $1\leq p < \infty$ and $\alpha$ ranges over bijections from $D_1$ to $D_2$.
\end{theorem}
\begin{theorem}
Given a persistence diagram $D$, the corresponding persistence surface $\rho_D:\mathbb{R}^2\to \mathbb{R}$ is defined as 
$$
\rho_D(x,y) = \sum_{u\in T(D)} f(u)\phi_u(x,y)
$$
where $f$ is a weighting function and $\phi_u$ is a Gaussian with mean $u$ and variance $\sigma^2$. Then the persistence Image vector is obtained by integrating this surface over a disceretized and relevant subdomain.  
\end{theorem}
For a more detailed explanation about the above definitions see~\cite{adams2017persistence}. It is proved that persistence image vector is stable under $1$-Wasserstein metric.

\subsection{PIV of Sentences from Sentence Cloud}
In our analysis, we consider a public speaking as a collection of finite number of sentences $\{s_i\}_{i=1}^N$. Using the corresponding sentence embeddings in $\{v_i\}_{i=1}^N\in \mathbb{R}^d$, we represent each speaker's talk as a collection of points in $\mathbb{R}^d$. We call this collection the \emph{sentence cloud} of a speaker. After that we use TDA to extract topological features of this sentence cloud and train our model with an additional input of persistence image vectors. For creating the vietoris-Rips complex from the sentence cloud, we used cosine dissimilarity metric defined as 
$$
d(s_1,s_2)= 1-\left(\frac{v_1\cdot v_2}{\norm{v_1}.\norm{v_2}}\right)/\pi
$$
where vector $v_i\in \mathbb{R}^d$ corresponds to the embedding of sentence $s_i$.
\section{Data}
The data used in our model was collected by crawling the \url{ted.com} website which contains more than 2300 talks given at public speaking setup. In the raw data we have the total number of views, the transcript used by the speaker, and the rating that the viewers have given to the talk. The rating contains 14 categories: \textsc{beautiful}, \textsc{confusing}, \textsc{courageous}, \textsc{fascinating}, \textsc{funny}, \textsc{informative}, \textsc{ingenious}, \textsc{inspiring}, \textsc{jaw-dropping}, \textsc{long-winded}, \textsc{obnoxious}, \textsc{ok}, \textsc{persuasive}, and \textsc{unconvincing}. For our analysis we used the transcript and the ratings given by the viewers for that talk. 
We normalize each rating labels by dividing it with the total view count and tranform each of the rating to binary labels with respect to the corresponding median (similar to  ~\citet{acharyya2019fairyted}). We train our classification model to predict each of this 14 rating labels.  

\section{Experiment}
\begin{figure}
\centering
\includegraphics[width=1.0\linewidth]{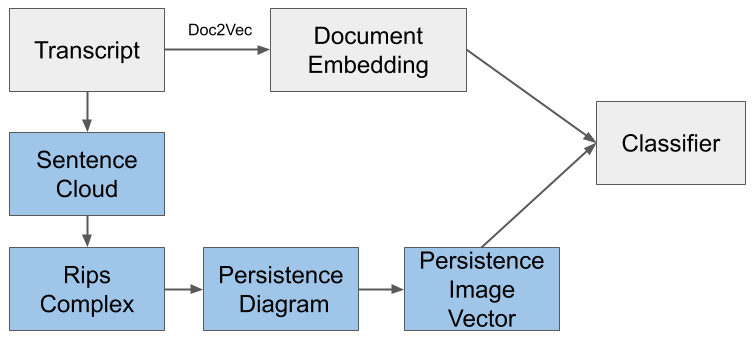}
\caption{Experiment Pipeline}
\label{fig:pipeline}
\end{figure}

In our experiment we train our model to predict the rating label of the ted-talks using the vector representation of the transcript of each talk. In the baseline method, we use only doc2vec representation of the transcripts for classification. Then we incorporate the topological features of the transcripts by calculating the PIV of the transcripts in the classifier.

The pipeline of our experiments is shown in Figure~\ref{fig:pipeline}. The gray part of the pipeline shows our baseline experiment and the blue part shows our method where we incorporate additional topological signature in the classifier.

\subsection{Vector Representation of Transcripts}
We utilize doc2vec implementation of \texttt{Gensim} package~\cite{le2014distributed} to create a vector representation of transcript in a high dimensional space $\mathbb{R}^{200}$.  As a baseline, we train our model to predict the rating of a public speaker on this doc2vec representation of the transcript. Then we experiment with various classification methods such as Support Vector Machine (SVM), Logistic Regression (LR), Multi Layer Perceptron (MLP) to train our classifer that predicts each of the 14 rating labels.

\subsection{Vietoris-Rips Complex of Sentence Embeddings}
For each talk we find the embedding of the sentences found in the transcript by using 4 state of the art embedding methods. 

    1) \textbf{BERT}~\cite{devlin2018bert}: BERT has recently achieved state of the arts results in various benchmark NLP tasks including question answering, natural language inference etc. We use `bert-as-service'~\cite{xiao2018bertservice} to get our embedding.
    
    2) \textbf{USE}~\cite{cer2018universal} Universal sentence encoder specifically targets transfer learning for various NLP tasks. There are two variations of the pre-trained USE model. We use the transformer encoder version in our experiment.
    
    3) \textbf{InferSent}~\cite{conneau-EtAl:2017:EMNLP2017} This model was trained on a natural language inference task and it was shown experimentally that this transfers well to other standard NLP tasks.
    
    4) \textbf{GenSen}~\cite{subramanian2018learning} This model is trained to learn general purpose, fixed-length representations of sentences via multi-task training which are useful for transfer and low-resource learning.
\
After that we use \texttt{ripser} package~\cite{ctralie2018ripser} to create the Vietoris-Rips complex for each talk by these sentence embeddings. This space contains the topological information about a talk which we extract by calculating the persistence diagram. Finally we create PIV of these diagrams. 
\subsection{PIV of a Talk} After we get the sentence cloud of a talk,  we use \texttt{persim} package~\cite{scikittda2019} to calculate the persistence image vector of each talk. Note that topological features appears and disappears in every dimension of the space, however for our analysis we use the 1-dimensional persistence diagram (see ~\citet{adams2017persistence} for detail). We vectorize the persistence image with  30$\times$30 pixels and use 0.01 as variance for the Gaussian (see section~\ref{piv}).  
\subsection{Rating Prediction of a Talk}
In our final step, we train different classification models such as SVM, LR, MLP; with and without the topological signatures found in the persistence image vectors to predict each of the 14 rating labels of a talk.
\section{Results}
\begin{figure*}
\centering
\includegraphics[width=0.95\linewidth]{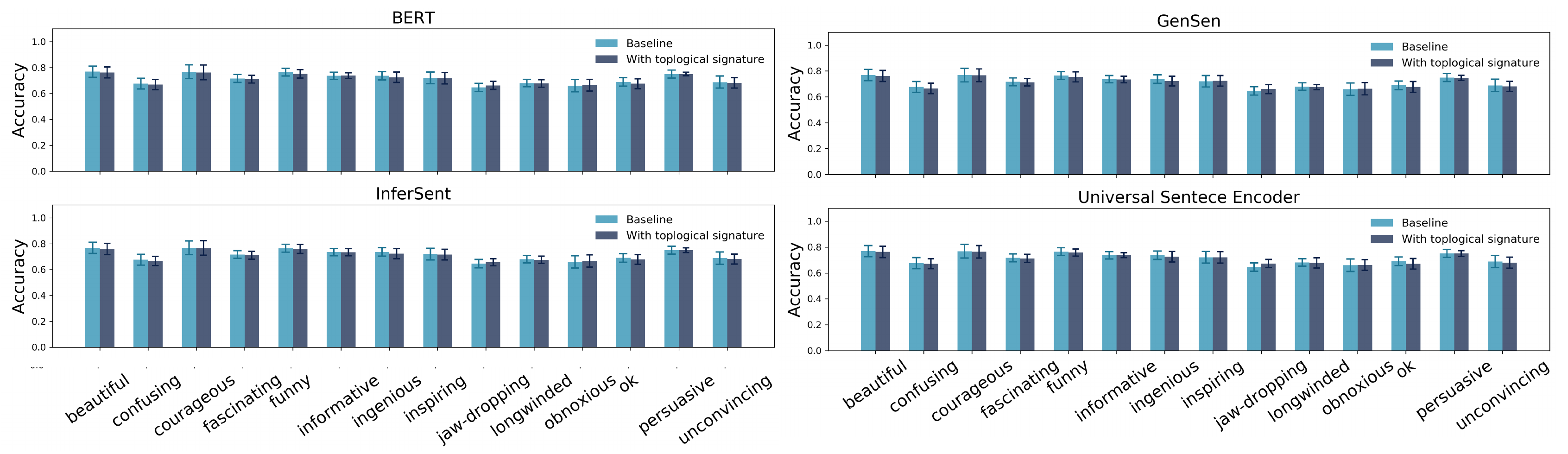}
\caption{10-fold cross validation accuracy using SVM classifier for different embedding methods}
\label{fig:accuracy_comparison}
\end{figure*}

\begin{table*}[t]
    \caption{Comparison of average accuracy for different classifiers}
    \begin{center}
        
    \begin{tabular}{|c|c|c|c|c|c|c|}
\hline
          \textrm{Embedding} &  \textrm{SVM}  &  \textrm{SVMTop} &  \textrm{LR} &  \textrm{LRTop} &  \textrm{MLP} &  \textrm{MLPTop} \\
\hline
                      \textrm{BERT} &           0.715 &                             0.711 &          0.686 &                            0.685 &           0.682 &                             0.690 \\
                 \textrm{InferSent} &           0.715 &                             0.710 &          0.686 &                            0.680 &           0.682 &                             0.688 \\
                    \textrm{GenSen} &           0.715 &                             0.711 &          0.686 &                            0.681 &           0.682 &                             0.689 \\
                    \textrm{USE} &           0.715 &                             0.712 &          0.686 &                            0.686 &           0.682 &                             0.689 \\
\hline
\end{tabular}
    \end{center}
    \label{tab:my_label}
\end{table*}
Among all the models we have used in our experiment, we found that SVM provides the best accuracy. Therefore we are showing the results of SVM in Figure~\ref{fig:accuracy_comparison}. We observe that our model never outperform the baseline when we include the topological attributes except for two rating category: \textsc{jaw-dropping} and \textsc{obnoxious}. The 10-fold cross validation accuracy remains almost same even after we add the topological features as additional inputs to the SVM model. In some cases, the accuracy becomes slightly worse. As indicated by~\cite{michel2017does}, it is possible that persistence image vector is not capturing any discriminitive feature because of the sparsity and small size of the sentence cloud. However, our work is different from \cite{michel2017does} because they used word level information where we used the sentence cloud of each talk. To check the robustness of this result, we also train LR and MLP as our rating predictor and compare the results with and without topological features obtained by the PIV. Table~\ref{tab:my_label} shows the average accuracy for each of the 14 rating labels across all models and embedding techniques. There is another interesting observation from the table~\ref{tab:my_label}. Although MLP model under-performs comparing to SVM and LR, it's average accuracy is improved when the topological features are added as an additional input. This could be a promising indication that if we could train neural network with a huge amount of data, the performance could benefit from the topological features prevalent in the high dimensional data points. 
\section{Conclusion}
TDA has a solid mathematical background and has recently been applied to a wide range of ML tasks with promising results~\cite{ferri2019topology}. In this paper we introduce the idea of using persistence image vector in the setting of  NLP task such as predicting the rating of public speaking. To the best of our knowledge, our attempt is the first of its kind to apply TDA to analyse a public speaking dataset. The shape of storytelling has been a widely explored topic but we propose a novel idea of considering sentence cloud to figure out the topological shape of public speaking. Although incorporation of the topological signature did not yield any significant improvement compared to the baseline in the classification task we proposed, it provides the NLP community a new pathway on using similar TDA techniques in other machine learning tasks. One of the difficulty in using TDA is choosing an appropriate metric and we believe that further examination of the use of PIV and similar TDA representations in different NLP tasks could be interesting direction in finding structure or geometric shapes of texts. One interesting but possibly difficult research question would be if we can develop any newer sentence embedding which can incorporate meaningful topological features of the sentence cloud of a speaker.

 \bibliography{acl2020}
 \bibliographystyle{acl_natbib}

\end{document}